\pdfoutput=1
\documentclass[11pt]{article}

\usepackage[]{acl} %review

\usepackage{times}
\usepackage{latexsym}

\usepackage[T1]{fontenc}
\usepackage[utf8]{inputenc}

\usepackage{microtype}

\usepackage{here}
\usepackage{latexsym}
\usepackage{booktabs}
\usepackage{stfloats}
\usepackage{subcaption}
\usepackage[T5,T1]{fontenc}

\usepackage{amsmath}
\usepackage{amsfonts}
\usepackage{cleveref}
\usepackage{float}

\usepackage{todonotes}
\usepackage{cancel}
\usepackage{xspace}
\usepackage{amsthm}
\usepackage{amssymb}
\usepackage{comment}
\usepackage{adjustbox}
\usepackage{tipa}
\usepackage{enumitem}
\usepackage{textcomp}

\usepackage{booktabs}
\usepackage{multirow}

\title{Analyzing the Representational Geometry of Acoustic Word Embeddings}

\author{Badr M. Abdullah  \and  Dietrich Klakow \\  Language Science and Technology (LST), Saarland University, Germany  \\ Saarland Informatics Campus \\ %$\dagger$ Saarland Informatics Campus \textsuperscript{1,2} \\ Spoken Language Systems (LSV) \\ 
       \normalsize{\textsf{\{ babdullah | dietrich \}@lsv.uni-saarland.de}} 
}

\begin{document}
\maketitle
\begin{abstract}
Acoustic word embeddings (AWEs) are vector representations such that different acoustic exemplars of the same word are projected nearby in the embedding space. 
In addition to their use in speech technology applications such as spoken term discovery and keyword spotting, AWE models have been adopted as models of spoken-word processing in several cognitively motivated studies and have been shown to exhibit human-like performance in some auditory processing tasks. 
Nevertheless, the representational geometry of AWEs remains an under-explored topic that has not been studied in the literature.
In this paper, we take a closer analytical look at AWEs learned from English speech and study how the choice of the learning objective and the architecture shapes their representational profile. 
To this end, we employ a set of analytic techniques from machine learning and neuroscience in three different analyses: embedding space uniformity, word discriminability, and  representational consistency. 
Our main findings highlight the prominent role of the learning objective on shaping the representation profile compared to the model architecture.

\end{abstract}

\section{Introduction}

Due to their ubiquity, word embeddings are nowadays a central component in natural language processing (NLP). 
Inducing word embeddings from text yields representations such that words occurring in similar contexts are nearby in the vector space \cite{mikolov2013distributed, pennington2014glove}. Therefore, the representational geometry of text-based word embeddings captures  lexical similarity and  semantic relatedness at multiple levels of granularity.
Word embeddings, and their underlying distributional  semantic  models, have also been adopted as models of human semantic memory in cognitive science research \citep[]{pereira2016comparative, Nematzadeh2017EvaluatingVM,
grand2022semantic}.  

\begin{figure}[t]
    \centering
    \includegraphics[width=0.5\textwidth]{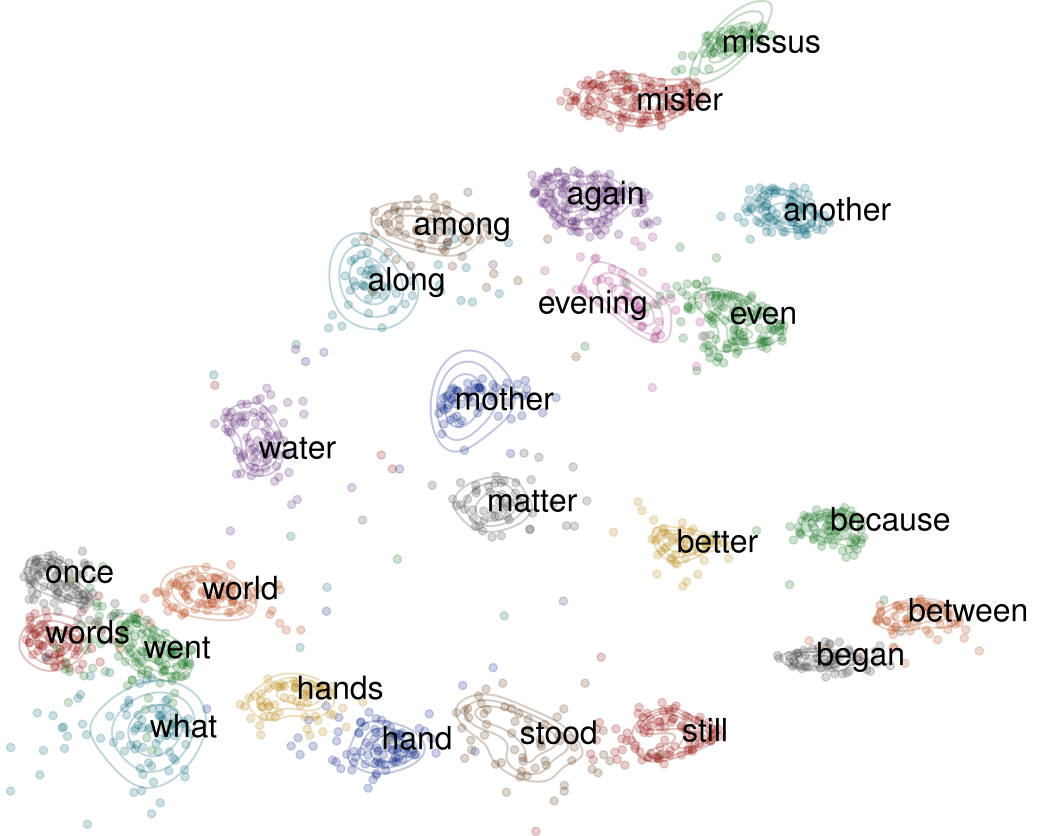}
    \caption{UMAP projection \citep{mcinnes2018umap} of a sample of acoustic word embeddings (AWEs) produced by a correspondence autoencoder (CAE) model trained on English read speech. AWE models project different exemplars of the same word type closer in the embedding space while abstracting away from speaker and context variability. }
    \label{fig:intro}
\end{figure}

% As a result, word embeddings have been shown to correlate with human judgments of lexical similarity and semantic relatedness. 
% Moreover, word embeddings have been adopted as models of the human lexicon since they simulate the mental lexical-semantic representations.
% models of lexical memory and processing 
In the speech processing domain, researchers have independently developed representations of acoustic segments that correspond to linguistic units \citep[\textit{inter alia}]{levin2013fixed,bengio2014word,kamper2016deep,settle2016discriminative}. 
A notable example of such representations are acoustic word embeddings (AWEs)---vector representations that encode the sound structure of words, not their semantic and syntactic structure---see Fig.~\ref{fig:intro}. 
AWEs support voice-based speech technology applications such as query-by-example spoken term discovery ~\cite{zhang2009unsupervised,jansen2012indexing,metze2013spoken} and keyword spotting \cite{myers1980investigation, rohlicek1995word}. 
In addition, AWEs can be leveraged to facilitate access to speech recordings of endangered spoken languages that might lack standardized writing systems \cite{bird2021sparse, san2021leveraging}

However, there are fundamental differences between text-based and speech-based word embeddings  that have to do with the degree of variability between the two modalities.  
Contrary to written words which have context-invariant orthographic realizations,\footnote{although some orthographic variation exists in informal, user-generated text such as tweets.} spoken words are notoriously variable.
The underlying sources of variability in speech include speaker-related factors such as vocal tract shape, gender, age, and dialect. 
In addition, two acoustic instances, or exemplars, of the same word will vary in different phonological and semantic contexts even if they are produced by the same speaker \cite{jurafsky2003probabilistic}. 
Therefore, acoustic word embeddings are not static, but have to be computed ``on the fly'' given a speech segment as input.
Models of AWEs need to abstract away from speaker and context variability to project different acoustic exemplars of the same word onto (ideally) the same point of the embedding space.

Nevertheless, AWEs have not yet been extensively studied in the literature from a neural network interpretability point of view.
We are only aware of a few prior efforts in this direction that have either analyzed the representational geometry of AWEs from a cognitively motivated angle \cite{matusevych+etal_baics20, Abdullah2021DoAW} or from a cross-linguistic perspective \cite{abdullah-etal-2021-familiar}.
In this paper, we make a contribution in this direction and  use analytic techniques from machine learning and neuroscience in three different analytic studies: embedding space uniformity (\S\ref{section_isotropy}), word discriminability  (\S\ref{discriminability}), and representational consistency (\S\ref{representational_consistency}).

\section{Acoustic Word Embedding Models}
Given an acoustic signal that corresponds to a spoken word represented as a temporal sequence of $T$ acoustic feature vectors, i.e., ${\boldsymbol{a}} = (\boldsymbol{a}^1, \boldsymbol{a}^2, ..., \boldsymbol{a}^T)$, the goal of an AWE model is to transform $\boldsymbol{a}$ into a fixed-dimensionality vector representation $\mathbf{e}$. 
Due to the variability in speech production (i.e., speech rate, emotional state, etc), the length of the acoustic segment $T$ varies between different exemplars, or instances, of the same word type. 
Therefore, this task is modeled as a mapping $\mathcal{F}: \mathcal{A} \xrightarrow[]{} \mathbb{R}^D$, where $\mathcal{A}$ is the (continuous) space of acoustic sequences and $D$ is the dimensionality of the embedding. 
%, and  $\boldsymbol{\theta}$ are the parameters of the function. 
Formally, transforming a variable-length acoustic input into a $D$-dimensional AWE is described as 
\begin{equation}
\mathbf{e} = \mathcal{F}(\boldsymbol{a}; \boldsymbol{\theta}_\mathcal{F}) \in \mathbb{R}^D
\end{equation}
where $\boldsymbol{\theta}_\mathcal{F}$ are the parameters of the encoder function $\mathcal{F}$. 
In a supervised setting of training AWE models, one assumes a dataset  $\mathcal{D} = \{(\boldsymbol{a}_1, w_1), (\boldsymbol{a}_2, w_2), \dots, (\boldsymbol{a}_N, w_N)\}$ of $N$ spoken word instances where $w_{i}$ is the word type, or lexical category, of the $i$th acoustic sample. 
In this paper, we experiment with two architectural choices---recurrent and convolutional---and employ four different learning objectives for training AWE models from the literature. 
Next, we formally describe each of the objectives.

\subsection{Correspondence Autoencoder}

In the correspondence autoencoder (CAE) \cite{kamper2019truly}, each training acoustic word sample ${\boldsymbol{a}}$ is paired with another sample that corresponds to the same word type ${\boldsymbol{a}_{+} = (\boldsymbol{a}^{1}_{+}, \boldsymbol{a}^{2}_{+}, ..., \boldsymbol{a}_{+}^{S}})$. The acoustic encoder $\mathcal{F}$  takes  ${\boldsymbol{a}}$  as input and produces an embedding $\mathbf{e}$, which is then fed to an acoustic decoder $\mathcal{H}$ that aims to sequentially reconstruct the corresponding acoustic sequence $\boldsymbol{a}_{+}$---see Fig.~\ref{fig:all_models}(a). The objective is to minimize the $L_2$ distance at each timestep in the decoder, which is equivalent to 
\begin{equation}
    J =   \sum_{i=1}^{S}{\lVert \boldsymbol{a}^{i}_{+} - \mathcal{H}^i(\mathbf{e})\rVert_2}
\end{equation}
where $\boldsymbol{a}_{i}^{+}$ is the ground-truth acoustic feature vector at timestep $i$ and $\mathcal{H}_i(\mathbf{e})$ is the reconstructed  acoustic vector at timestep $i$ as a function of the embedding $\mathbf{e}$. Learning the correspondence between different acoustic realizations of the same word type seems to encourage the encoder to build up speaker-invariant word representations while preserving linguistically-relevant phonetic information \cite{matusevych2020analyzing}. 
When the target acoustic sequence to generate is the same as the input signal $\boldsymbol{a}$, this corresponds to a conventional autoencoder (AE) which we consider as one of our learning objectives in this paper.

\subsection{Phonologically Guided Encoder}
The phonologically guided encoder (PGE) is trained as component in a sequence-to-sequence model to map acoustics into phonology \cite{Abdullah2021DoAW}. 
Given the output of the encoder as an embedding $\mathbf{e}$, a phonological decoder $\mathcal{G}(.; \boldsymbol{\theta}_\mathcal{G})$ is trained to decode the corresponding phonological sequence $\boldsymbol{\varphi} = (\varphi^{1}, \dots, \varphi^{\tau})$ of the word-form ---see Fig.~\ref{fig:all_models}(b).
The objective is to minimize a categorical cross-entropy loss at each decoder timestep, which is equivalent to minimizing the term
\begin{equation}
    \begin{aligned}
        J & = - \sum_{({\boldsymbol{a}}_i, w_i) \in \mathcal{D}}  { \text{log } \mathbf{P}\big( \boldsymbol{\varphi}^{} | \mathbf{e}_i ; \boldsymbol{\theta}_\mathcal{G}\big)} \\
        & = - \sum_{({\boldsymbol{a}}_i, w_i) \in \mathcal{D}} \sum_{t = 1}^{\tau}{ \text{log } \mathbf{P}\big(\varphi^{t}  |  t,  \mathbf{e}_i ; \boldsymbol{\theta}_\mathcal{G}\big)}
    \end{aligned}% 1:\tau
\end{equation} %\boldsymbol{\varphi}_{<t} $\mathcal{F}(.)$ as a single vector
where $\mathbf{P}\big(\varphi^{t} | t, \mathbf{e}_i ; \boldsymbol{\theta}_\mathcal{G}\big)$ is the probability of the phoneme $\varphi^{t}$ at the $t$th timestep, conditioned on the previous phoneme sequence $\boldsymbol{\varphi}^{<t}$ and the AWE $\mathbf{e}$, and $\boldsymbol{\theta}_\mathcal{G}$ are the parameters of the decoder. 
The intuition of this learning objective is the following: although their acoustic realizations vary due to speaker and context variability, different exemplars of the same word category would have identical phonological sequences.
We thus expect the encoder to project exemplars of the same lexical category nearby in the embedding space while embedding similarity in the vector space should correlate with phonological similarity.
% Given an acoustic sequence ${\boldsymbol{a}}$ and its corresponding phonological sequence $\boldsymbol{\varphi} = (\varphi_1, \dots, \varphi_{\tau})$,\footnote{We use the phonetic transcription in IPA symbols as the phonological sequence $\boldsymbol{\varphi}$.} the acoustic encoder $\mathcal{F}$ is trained to take ${\boldsymbol{a}}$ as input and produce an AWE $\mathbf{e}$, which is then fed into a phonological decoder $\mathcal{G}$ whose goal is to generate the sequence $\boldsymbol{\varphi}$ (Fig.~\ref{fig:models}--a).  The objective is to minimize a categorical cross-entropy loss at each timestep in the decoder, which is equivalent to 
% \begin{equation}
%     J = -  \sum_{i=1}^{\tau}{ \text{log } \mathbf{P}_{\mathcal{G}}(\varphi_{i} | \boldsymbol{\varphi}_{<i}, \mathbf{e})}
% \end{equation}
% where $\mathbf{P}_{\mathcal{G}}$ is the probability of the phone $\varphi_{i}$ at timestep $i$, conditioned on the previous phone sequence $\boldsymbol{\varphi}_{<i}$ and the embedding $\mathbf{e}$. The intuition of this objective is that spoken-word segments of the same word type would have identical phonological sequences, thus they are expected to end up nearby in the embedding space.

% \boldsymbol{a}^{+} 
% \boldsymbol{a}

\subsection{Contrastive Siamese Encoder }
The contrastive siamese encoder (CSE) has been explored in the context of AWEs with both recurrent and convolutional architectures in several studies \cite{settle+livescu_slt16, kamper+etal_icassp16, jacobs2021acoustic}.  
Contrary to the previously described objectives, the CSE explicitly minimizes the distance between exemplar embeddings of the same word type---see Fig.~\ref{fig:all_models}(c). 
First, each acoustic word instance is paired with another instance of the same word type  $(\boldsymbol{a}, \boldsymbol{a}_+)$. 
Given their embeddings $(\mathbf{e}_a, \mathbf{e}_+)$, the objective is then to minimize a triplet margin loss
\begin{equation}
    J = \text{max} \big[0, m + d(\mathbf{e}_a, \mathbf{e}_+) - d(\mathbf{e}_a, \mathbf{e}_-) \big]
\end{equation}
Here, $d(., .)$ is the cosine distance and $\mathbf{e}_-$ is an AWE that corresponds to a different word type sampled from the mini-batch such that the term $d(\mathbf{e}_a, \mathbf{e}_-)$ is minimized. 
This objective clusters acoustic instances of the same word type closer in the embedding space while pushing away instances of other word types by a distance defined by the margin hyperparameter $m$. 

\section{Data, Setup, and Intrinsic Evaluation}
\subsection{Experimental Data}

The data in our study is drawn from the the LibriSpeech dataset which contains read speech recordings of American-English  \cite{panayotov2015librispeech}, which is a public dataset under the CC BY 4.0 license.
We sample 384 speakers from for training and 128 for evaluation---disjoint sets---and obtain word-aligned speech samples using the Montreal Forced Aligner \cite{mcauliffe2017montreal}. 
To make our models comparable with prior work, which has focused on AWEs for low-resource languages, we sample  $\sim39.4$k samples for training and $\sim9.7$k for evaluation.
The phonetic transcription for each word is produced using the online \textit{WebMaus} G2P tool \cite{strunk2014untrained}.  
Then, each acoustic segment is parametrized  as a sequence of 39-dimensional Mel-frequency spectral coefficients of 25ms frames with 15ms overlap---the conventional feature representation of speech in automatic speech recognition (ASR). 
It is worth pointing out that in this paper we consider each morphological variant of a lexeme  as a separate lexical category. 
For example, different inflections of the lexeme \textsc{make} such as \{\textsc{made}, \textsc{making}, \textsc{maker}, etc.\} represent different lexical categories, each with its own exemplars.

\subsection{Architectures, Hyperparameters, and Training Details}
\textbf{CNN Acoustic Encoder.} \hspace{0.15cm} We employ a 3-layer temporal convolutional network (1D-CNN) with 256, 384, and 512 filters and widths of 4, 8, and 16 for each layer and keep stride step at 1. Following each convolutional operation, we apply batch normalization, ReLU non-linearity, and dropout. We apply average pooling to downsample the representation at the end of the convolution block, then apply one non-linear layer with Tanh on the CNN output, which yields a 512-dimensional AWE.\vspace{0.1cm} %

% , with a probability tuned  {0.0, 0.2, 0.4} for each learning objective
% For each learning in our study, we experiment with {0, 1, 2} fully-connected layers on top of the convolutional block.
\noindent
\textbf{RNN Acoustic Encoder.} \hspace{0.15cm} We employ a 3-layer directional Gated Recurrent Unit (GRU) with a hidden state dimension of 512, then apply one non-linear layer with Tanh on the GRU output, which yields  a 512-dimensional AWE.
We apply layer-wise dropout with a probability of $0.1$. \vspace{0.1cm}

\noindent
\textbf{Phonological Decoder $\mathcal{G}(.; \boldsymbol{\theta}_{\mathcal{G}})$.} \hspace{0.1cm} We employ a 1-layer GRU of 512 units hidden state that takes the 512-dimensional AWE as the initial hidden state and decodes the corresponding phonological sequence without teacher forcing.  \vspace{0.1cm}

\noindent
\textbf{Acoustic Decoder $\mathcal{H}(.; \boldsymbol{\theta}_{\mathcal{H}})$.} \hspace{0.1cm} We employ a 1-layer GRU of 512 units hidden state that takes the 512-dimensional AWE as the initial hidden state and decodes the corresponding acoustic sequence with a teacher forcing ratio of 0.2.  \vspace{0.1cm}

% and experiment with {0, 1, 2} fully-connected layers on top of the recurrent block for each learning objective.
% Phonetic Decoder
% For the word-to-phonemes objective, we employ a 1-layer GRU with a 1024-dimensional hidden state, 128-dimensional phoneme embedding, and apply teacher forcing during training. The AWE from the acoustic encoder is used to initialize the first hidden state of the decoder. 
% Classification layer
% For the Phoneme n-gram detection objective, we attach a multi-class multi-label classification layer on the top of the AWE. We keep only the phoneme bigrams and trigrams that occur at least 200 times in the training data. This threshold is tuned on the validation set of the German data. 

\noindent
\textbf{Contrastive Loss.} \hspace{0.1cm} For the CSE, we experiment with different values of the margin hyperparameter $m= \{0.2, 0.3, 0.4, 0.5\}$, out of which $0.4$ yields the best performance on the validation set.
\vspace{0.1cm}

\noindent
\textbf{Training Details.} \hspace{0.15cm} All models in this study are randomly initialized with each parameter drawn uniformly from $[-0.05, 0.05]$. 
Then, each model is trained for 100 epochs with a batch size of 256  using the ADAM optimizer \cite{DBLP:journals/corr/KingmaB14} and an initial learning rate of 0.001. 
The learning rate is reduced by a factor of 0.5 if the mAP on the validation set does not improve for 10 epochs.
% The epoch with the best validation performance during training is used for evaluation on the test set. \vspace{0.1cm}

\noindent
\textbf{Implementation.} \hspace{0.15cm}  We build our models using PyTorch \cite{paszke2019pytorch}  and use FAISS \cite{JDH17} for efficient similarity search.
Our code is based on our prior work in building and analyzing AWEs \cite{Abdullah2021DoAW, abdullah-etal-2021-familiar}.

% See appendix \S\ref{app:arch_hyp}  for a detailed overview of the architectures of the model in this paper. 
% We will move this section back to the main body of the paper in the final camera-ready  version.

\begin{figure*}[t]
    \centering
    \includegraphics[width=0.99\textwidth]{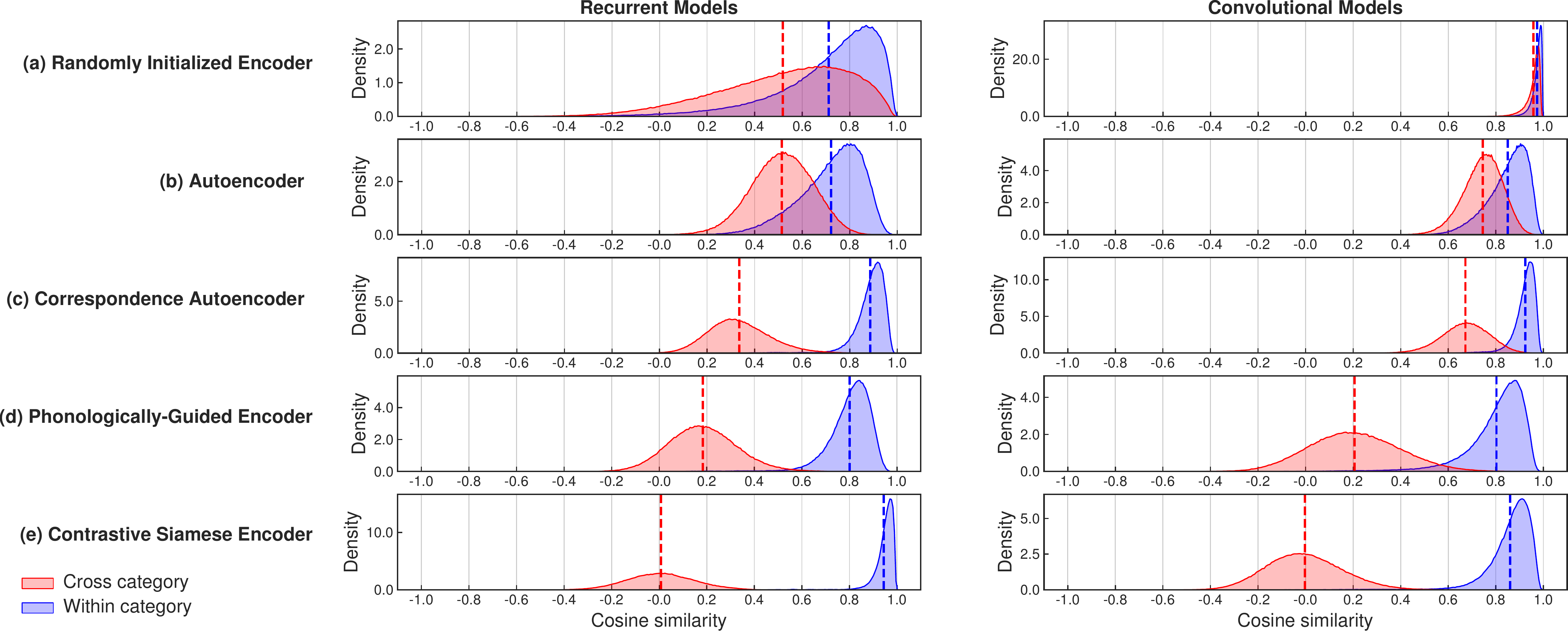}
    \caption{Distribution of cosine similarity scores of the different models for within category samples (i.e., exemplar pairs of the same word type) and cross-category samples (i.e., sample pairs that correspond to different word types). Each row in the figure corresponds to one learning objective and each column corresponds to one architecture. }
    \label{fig:cosine_sim_dist}
\end{figure*}

\subsection{Quantitative Evaluation}
We conduct an intrinsic evaluation for the AWEs to assess the performance of our models using the same-different acoustic word discrimination task with the mean average precision (mAP) metric \cite{carlin2011rapid, kamper2015unsupervised, settle+etal_icassp19, algayres20_interspeech}.  
% Prior work has shown that performance on this task positively correlates with improvement on downstream QbE speech search \cite{Jacobs2021MultilingualTO}.  
This task evaluates the ability of the model to determine whether two given speech segments correspond to the same word type---that is, whether or not two acoustic segments are exemplars of the same category.  
The results of the evaluation is shown in Fig.~\ref{fig:AWE_evaluation} in the appendix.
We observe that each recurrent encoder outperforms its convolutional counterpart within each objective. 
Moreover, the performance largely depends on the strength of the supervision signal where the contrastive encoders outperform other objectives that lack explicit loss to group exemplars of the same category closer in the embedding space.

% We evaluate the models using the standard intrinsic evaluation of AWEs: the same-different word discrimination task \cite{carlin2011rapid}. This task aims to assess the ability of a model to determine whether or not two given speech segments correspond to the same word type, which is quantified using a retrieval metric (mAP) reported in Table~\ref{tab:mAP}.

\section{Analysis 1: Embedding Space Uniformity}
\label{section_isotropy}
In our first analysis, we take a closer look at how uniform are representational spaces of AWE models by analyzing the distribution of cosine similarity for each model type and the degree to which the embeddings are isotropic.

\subsection{Distribution of Cosine Similarity}
One way of analyzing the geometry of representation spaces in the acoustic domain is by inspecting the similarity distributions of exemplars of the same lexical category (or word type) versus randomly sampled, cross-category exemplars.  
We perform this analysis on the training samples and depict the result in Fig.~\ref{fig:cosine_sim_dist}. 
We observe that the difference between the means of the within-category and those of cross-category distributions is largely dependent on the strength of the supervision signal with the randomly initialized encoders (RIE) having the smallest mean differences for both architectures. 
The contrastive encoders have the largest mean difference--with mean cross-category scores centered at the zero---which is intuitive given the explicit supervision signal they receive in grouping exemplars of the same category closer in the embedding space. 
One surprising observation is the behavior of the untrained convolutional encoder which gives cosine similarity scores very close to 1 for each input pair. 
In appendix~\ref{sec:appendix_cnn_cosine}, we demonstrate that this behavior is mainly caused by the unbounded activation function (i.e., ReLU) in the convolutional layers. 

\subsection{Degree of Isotropy}

Although inspecting the cosine similarity distributions is an insightful analysis, it does not enable us to make well-informed judgments about the uniformity of the representation spaces. Here, we ask two questions: (1) do AWE models utilize all dimensions of the vector space to represent the speech samples and separate the categories? and (2) how do architecture and learning objective affect the distributivity of information in the embedding space?
To answer these questions, we inspect the degree of isotropy in the representation spaces. 
An embedding space is said to be maximally isotropic if the variance is uniformly distributed across all dimensions.
%On the other hand, an embedding space is said to be minimally iostropic if the points only vary along a single dimension.
Prior work in NLP has found that semantic word embeddings tend to be anisotropic since they only utilize a few dimensions of the vector space---an effect that has been observed for word embeddings that are static \citep{mimno2017strange, mu2018all} as well as contextualized  \citep{ethayarajh-2019-contextual, cai2020isotropy, rudman-etal-2022-isoscore}. 
The degree of isotropy in acoustic embeddings, however, remains so far unknown.
\begin{figure}[t]
    \centering
    \includegraphics[width=0.5\textwidth]{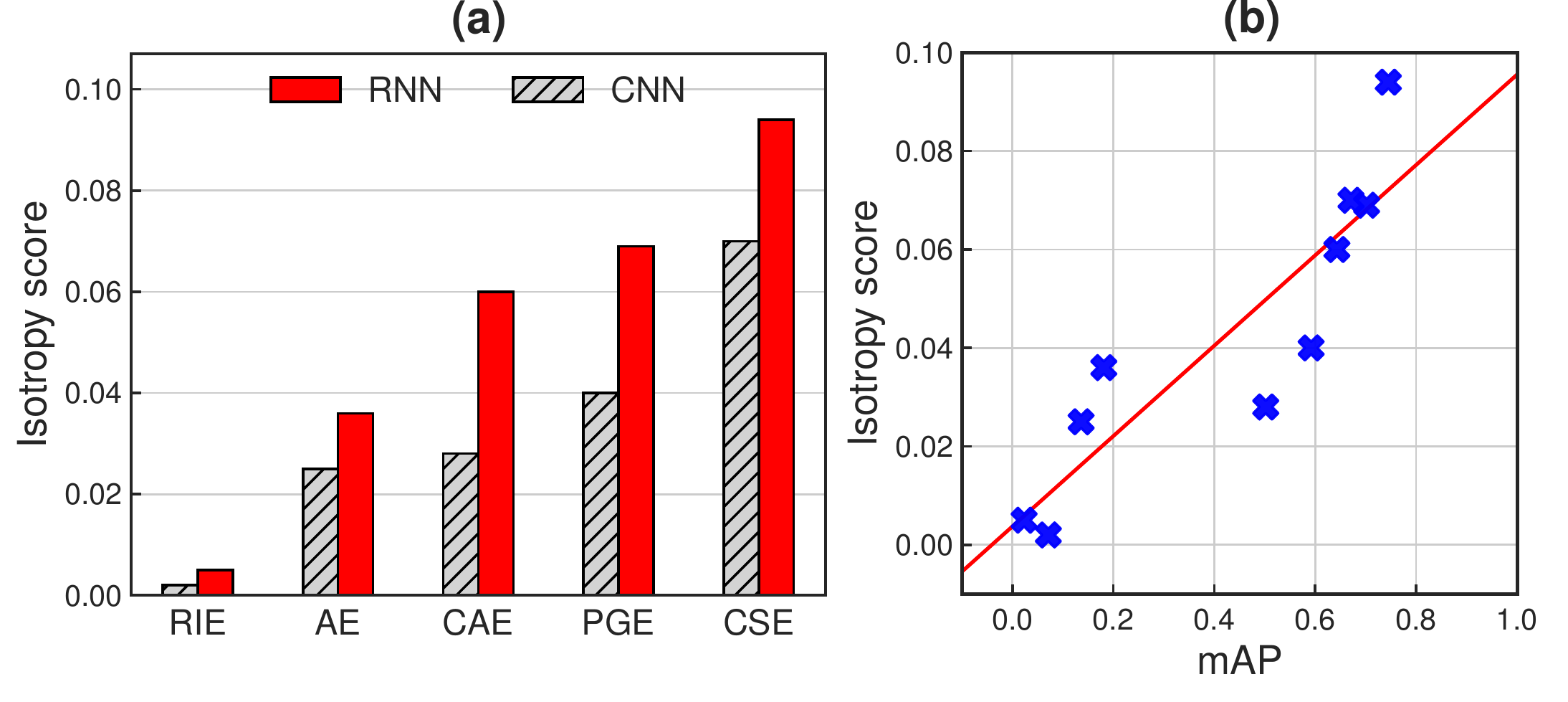}
    \caption{(a). The degree of isotropy of AWE for each model. (b) Correlation between the word discrimination performance measured by mAP and isotropy score (Pearson $r = 0.89, p <0.001$). }
    \label{fig:isotropy}
\end{figure}
% they occupy a ``narrow cone'' within the embedding space
To inspect the degree of isotropy of the AWE vector spaces, we use the IsoScore metric recently proposed by \citet{rudman-etal-2022-isoscore}, which is---to the best of our knowledge---the only metric in the literature that is grounded on the mathematical definition of isotropy. The IsoScore metric operates on the covariance matrix of the embedding dimensions and returns values between 0 (minimally isotropic) and 1 (maximally isotropic). 
We quantify the degree of isotropy using IsoScore for each model type and show the result in Fig.~\ref{fig:isotropy}(a). 
We observe that IsoScore returns values that are within the range [$0.002, 0.095$], which indicates that embedding spaces for all models tend towards being minimally isotropic. 
However, the embeddings of untrained, randomly initialized encoders (RIE) tend to be extremely anisotropic (i.e., IsoScore values close to 0). 
This observation suggests that the anisotropic space does not ``emerge'' during the model training but rather that it is an inherent property of the encoder architecture. 
We are not aware of prior work in NLP that has studied the degree of isotropy in untrained NLP models to investigate whether anisotropic spaces are an emergent or inherent feature. 
In our case, training with a learning objective that encourages the model to separate word categories moves the representation space more towards utilizing more dimensions, therefore resulting in a higher degree of isotropy.
Moreover, recurrent encoders tend to be more isotropic than their convolutional counterparts within the same learning objective. 

Despite the tendency of all models to be anisotropic, we find a strong positive correlation between the degree of isotropy and the performance on word discrimination---see Fig.~\ref{fig:isotropy}(b).
That is, the more dimensions the model utilizes in the representation space, the better it performs on the intrinsic evaluation task. 

\section{Analysis 2: Word Discriminability}
\label{discriminability}
Ideally, AWE models should project exemplars of the same word category onto the same point in the embedding space. 
However, there are no strong constraints during training to encourage maximal separability between different word categories. 
In this analysis, we seek to answer two questions: (1) how well-separated are the word categories of the training samples? and (2) to what degree do lexical properties predict the discriminability of word categories?

%such as category frequency and acoustic distinctiveness 

\subsection{Category Discriminability Index}
\begin{figure}[t]
    \centering
    \includegraphics[width=0.5\textwidth]{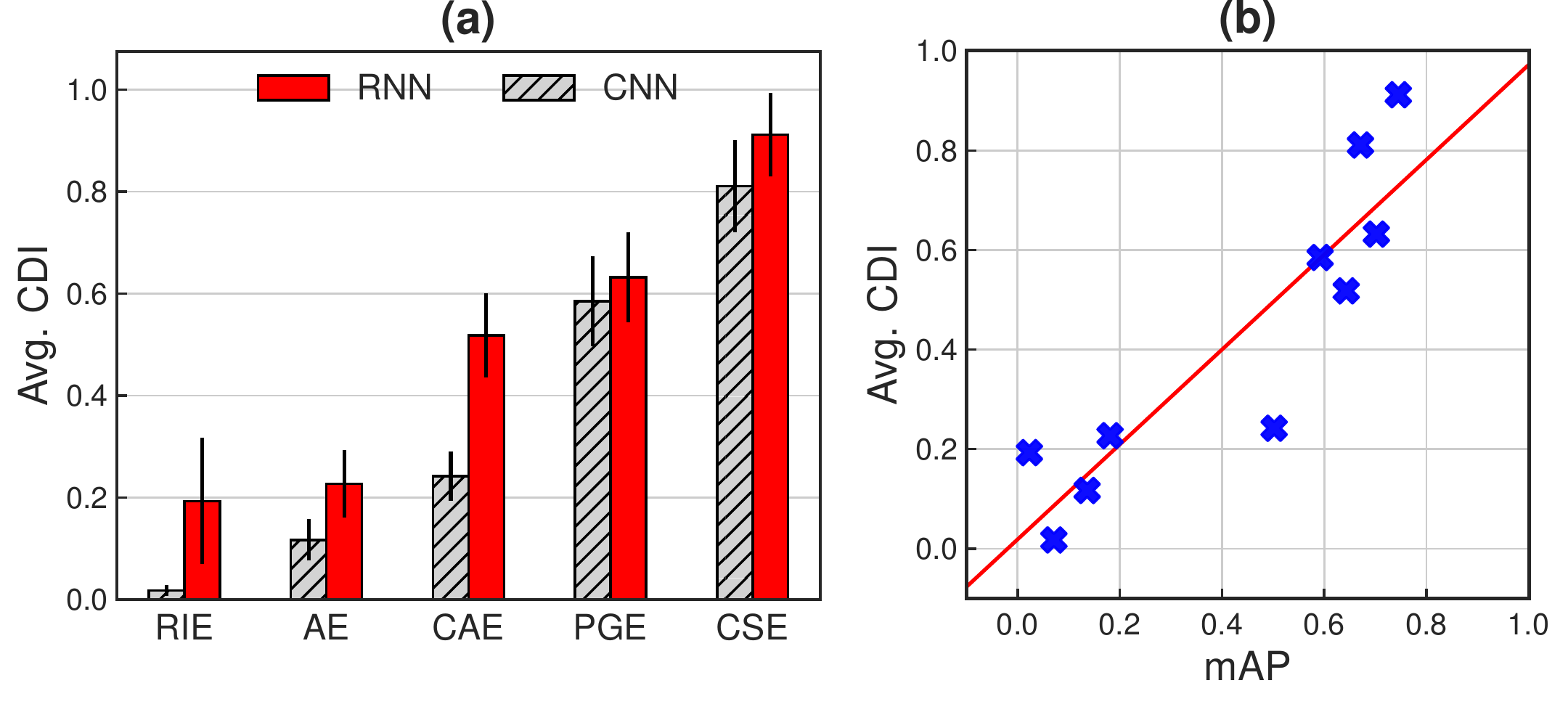}
    \caption{Averaged Category Discriminability Index (CDI) for each AWE model with error bars showing standard deviation over word categories. (b) Correlation between the word discrimination performance measured by mAP and averaged CDI (Pearson $r = 0.90, p <0.001$).} 
    \label{fig:CDI}
\end{figure}
In order to investigate the geometric density of each word category in the representation space, we need to measure within-category compactness and cross-category separability. 
Inspired by the exemplar discriminability index proposed in the neuroscience literature \cite{nili2020inferring}, we define category discriminability index (CDI) as a metric that operates on within-category and cross-category distances. 
If we consider each lexical category in the training set as a set of its exemplar embeddings $\mathcal{C} = \{\mathbf{e}_1, \dots, \mathbf{e}_{|\mathcal{C}|}\}$, CDI is defined for a single category $\mathcal{C}$ as  
\begin{multline}
\text{CDI}(\mathcal{C}) =  \frac{1}{|\mathcal{C}|}  \sum_{\forall \mathbf{e}_i \in \mathcal{C}} 
    \\ \bigl( \sum_{\forall \mathbf{e}_j \sim  \mathcal{C} | j \neq i}  {d(\mathbf{e}_i, \tilde{\mathbf{e}}_j)}   
  - d(\mathbf{e}_i, \mathbf{e}_j) \bigr)
\end{multline}
% \begin{multline}
% \text{CDI}(\mathcal{C}) = \\ \frac{1}{|\mathcal{C}|}  \sum_{\forall(\mathbf{e}_i, \mathbf{e}_j) \in \mathcal{C} | i \neq j} 
%   \Bigl( {d(\mathbf{e}_i, \tilde{\mathbf{e}}_k)}   
%   - d(\mathbf{e}_i, \mathbf{e}_j) \Bigr)
% \end{multline}
% % \\ \sum_{\mathbf{e}_k \sim  \mathcal{L} \setminus \mathcal{C}}
% d(\mathbf{e}_i, \mathbf{e}_k)
where $d(.,.)$ is the cosine distance and $\mathbf{e}_j$ is a within-category sample while $\tilde{\mathbf{e}}_j$ is an embedding sampled from a different category.  % \sim  \mathcal{L} \setminus \mathcal{C}
If we normalize the embeddings, CDI $\in [-1, 1]$ with values closer to $1$ indicating higher word discriminability.
We compute CDI for each word category in the training set and take the average over categories to estimate how well the categories are separated in the embedding space of each model type. 
The result of this analysis is shown in Fig.~\ref{fig:CDI}(a). 
For each learning objective, we observe that word discriminability is higher in the recurrent encoders compared to their convolutional counterparts. 
Besides that, the contrastive objective yields encoders with a higher word discriminability index regardless of the architecture type---recurrent vs. convolutional.  
Furthermore, we report a strong positive correlation between average CDI and the performance on the evaluation task---see Fig.~\ref{fig:CDI}(b), indicating that word discrimination performance on future, held-out samples can be predicted based on the CDI computed on the training samples. 

\subsection{Effect of Frequency and Distinctiveness}
The CDI quantifies the separability and compactness for each lexical category in the representation space. 
Next, we aim to identify the factors that could make a lexical category compact and well-separable. 

In this analysis, we study the effect of two lexical properties that could be quantified in a data-driven approach:  word frequency and acoustic distinctiveness. 
Our initial hypothesis is that a word category with many training exemplars becomes more discriminable in the embedding space as the repeated exposure to samples of various degrees of variability should enable the model to learn compact and precise representation for categories with high frequency.  
Also, words that are acoustically distinct have fewer competitors in the perceptual space, thus they should be more separable than words with many phonological neighbours that sound similar. 
Therefore, we expect word acoustic distinctiveness (WAD) to positively correlate with CDI. 
In this analysis, we operationalize WAD using two metrics: word length (i.e., the number of phonemes) and phonological distinctiveness. 
Word length contributes to WAD since word formation in natural languages is a combinatorial process.
That is, increasing the number of phonemes in a word-form decreases the likelihood of encountering a similarly sounding word-form which makes it less confusable. 
However, the word formation process is governed by language-specific phonotactic rules which makes some sound combinations more probable than others. 
To capture the probabilistic nature of sound sequences, we employ  phonological information content (PIC), an information-theoretic metric that estimates WAD based on its phoneme-to-phoneme transition probabilities \cite{Meylan2017WordF}. 
Given a word-form as a sequence of phonemes $\boldsymbol{\varphi} = (\varphi_1, \dots, \varphi_\tau)$, PIC is defined as 
\begin{equation}
\text{PIC}(\boldsymbol{\varphi}) = - \sum_{i=1}^{\tau}{\log p_{\boldsymbol{\theta}} (\varphi_i | \boldsymbol{\varphi}_{<i})}
\end{equation}
where $p_{\boldsymbol{\theta}}$ is a probabilistic phoneme-level language model (PLM). We estimate $p_{\boldsymbol{\theta}}$ using a trigram PLM with the counts of the phonemes in the training word categories. 
Higher values of PIC indicate less probable phoneme sequences thus more distinct word-forms. 
Note that PIC is not length normalized and therefore shorter words tend to have lower PIC. 

Next, we conduct a correlation analysis between word CDI and the three lexical predictors: frequency, length, and PIC. 
The result of this analysis is shown in Table~\ref{tab:CDI_corr}. 
Surprisingly, our correlation analysis shows that lexical frequency is a poor predictor of CDI. 
Although in five out of eight models the frequency positively correlates with CDI, the correlation is rather weak.
However, measures of acoustic distinctiveness have a stronger correlation with CDI compared to frequency, and the strength of the correlation is more noticeable in all decoding-based models---except the convolutional PGE---compared to contrastive models.
We also find it surprising that PIC is not a better predictor of CDI than word length.
However, it has been shown in a related work that autoencoder-based AWEs encode duration as an acoustic feature \cite{matusevych2021phonetic}. 
Taken together with our findings, this suggests that the models exploit and rely on acoustic word length as a feature to discriminate between the lexical categories. 
Arguably, word length is a more accessible feature to learn from the acoustic signal compared to structural phonological regularities in the training data.

% length offers far more explanatory for word discrimination 

% Please add the following required packages to your document preamble:
% \usepackage{booktabs}
% \usepackage{multirow}
\begin{table}[t]
\centering
\scalebox{0.85}{
    \begin{tabular}{@{}ccccc@{}}
    \toprule
    Objective                 & Arch. & Frequency & Length & PIC    \\ \midrule
    \multirow{2}{*}{AE}  & \textsc{cnn}   & -0.081†   & 0.315† & 0.263† \\
                         & \textsc{rnn}   & -0.087†   & 0.357† & 0.306† \\ \midrule
    \multirow{2}{*}{CAE} & \textsc{cnn}   & 0.021     & 0.376† & 0.274† \\
                         & \textsc{rnn}   & 0.077†    & 0.447† & 0.359† \\ \midrule
    \multirow{2}{*}{PGE} & \textsc{cnn}   & 0.035     & 0.039* & -0.011 \\
                         & \textsc{rnn}   & -0.043*   & 0.325† & 0.263† \\ \midrule
    \multirow{2}{*}{CSE} & \textsc{cnn}   & 0.131†    & 0.075† & 0.031  \\
                         & \textsc{rnn}   & 0.109†    & 0.100† & 0.030  \\ \bottomrule
    \end{tabular}
}
\caption{Pearson correlation ($r$) between word category discriminability index (CDI) and three lexical properties: frequency, length, and phonological information content (PIC). Statistical significance is marked with *  and † for $p < 0.05$ and $p < 0.001$, respectively.}
\label{tab:CDI_corr}
\end{table}
\section{Analysis 3: Network Representational Consistency}
\label{representational_consistency}
Suppose we train two instances of the same architecture and learning objective on the same training samples, but each with different random initializations. 
Do these two neural network instances exhibit differences in their representational geometries?
In this section, we shed light on the representational discrepancies caused by different initializations.
In other words, we are interested in quantifying the degree to which variability in the initial conditions affects the way two models separate the same set of speech samples. 

\subsection{Performance Stability}
First, we quantify the effect of the initial weights on the evaluation task performance.
To this end, we train six model instances---in identical setup but with different initializations---for each architecture and each learning objective, which yields 48 model instances in total (6 $\times$ 4 RNN runs and 6 $\times$ 4 CNN runs). 
We evaluate each model instance on the acoustic word discrimination task while observing the result variation per model type. 
The result of the performance stability analysis is shown in Table~\ref{tab:mAP_performance} in Appendix ~\ref{sec:appendix_RC}. 
We observe that  all instances have converged and the performance is fairly stable across different runs. 

\subsection{Representational Discrepancies}

Our previous performance stability analysis has demonstrated that different DNN instances exhibit only trivial quantitative differences.  
However, a stable performance on the evaluation task does not entail an identical representational geometry across different instances.
That is, two network instances could have have an identical performance on the evaluation task while each having a distinct representational geometry. 
To closely investigate representational discrepancies between network instances, we employ the representational consistency (RC) analysis \cite{mehrer2020individual}, which is a neuroscience-inspired technique based on the representational similarity analysis (RSA) framework \cite{kriegeskorte2008representational}.
For our analysis, we operationalize the RC using linear Centered Kernal Alignment (CKA) as a representational similarity measure of two views of the same input samples \cite{kornblith2019similarity}. 
CKA abstracts away from the embeddings themselves and operates on pairwise distances between the sample representations. 
Concretely, given $K$ spoken-word samples  $\boldsymbol{a}_{1}^{K} = \{\boldsymbol{a}_1, \dots, \boldsymbol{a}_K\}$, we embed the samples using two encoder instances  to obtain two different views of the samples $\mathbf{X} \in \mathbb{R}^{K \times D}$ and $\mathbf{Y} \in \mathbb{R}^{K \times D}$. 
Then, each view matrix is multiplied by a centering matrix $\mathbf{H} = \mathbf{I}_K - \mathbf{1}_K/K$ to make each column's mean equal to zero and obtain centered second moment matrices as 
\begin{align}
\begin{split}
 & \mathbf{G}_\mathbf{X} =  \mathbf{H}  \mathbf{X}  \mathbf{X}^\top  \mathbf{H}^\top /D, \\  
    & \mathbf{G}_\mathbf{Y} =  \mathbf{H}  \mathbf{Y}  \mathbf{Y}^\top  \mathbf{H}^\top /D  
\end{split}
\end{align}
% \begin{equation}
% \mathbf{G}_\mathbf{X} =  \mathbf{H}  \mathbf{X}  \mathbf{X}^\top  \mathbf{H}^\top /D,  
% \mathbf{G}_\mathbf{Y} =  \mathbf{H}  \mathbf{Y}  \mathbf{Y}^\top  \mathbf{H}^\top /D
% \end{equation}
Then, the representational similarity of the two views is computed using CKA as 
\begin{equation}
\text{CKA}(\mathbf{X}, \mathbf{Y}) = \frac{\langle \text{vec}( \mathbf{G}_\mathbf{X}), \text{vec}( \mathbf{G}_\mathbf{Y}) \rangle}{||  \mathbf{G}_\mathbf{X} ||_F ||  \mathbf{G}_\mathbf{Y}||_F}
\end{equation}
% \begin{equation}
% \text{CKA}(\mathbf{X}, \mathbf{Y}) = \frac{\langle \text{vec}(\mathbf{X} \mathbf{X}^\top), \text{vec}(\mathbf{Y} \mathbf{Y}^\top) \rangle}{|| \mathbf{X} \mathbf{X}^\top ||_F || \mathbf{Y} \mathbf{Y}^\top||_F}
% \end{equation}
where $\text{vec}(.)$ is the vector-reshaped matrix, $\langle ., .\rangle$ is the inner product, and $||.||_F$ is the Frobenius norm to ensure that CKA $\in [0, 1]$ where values close to 1 indicate that the two instances are highly consistent, while values close to 0 indicate low consistency.

\begin{figure}[t]
    \centering
    \includegraphics[width=0.45\textwidth]{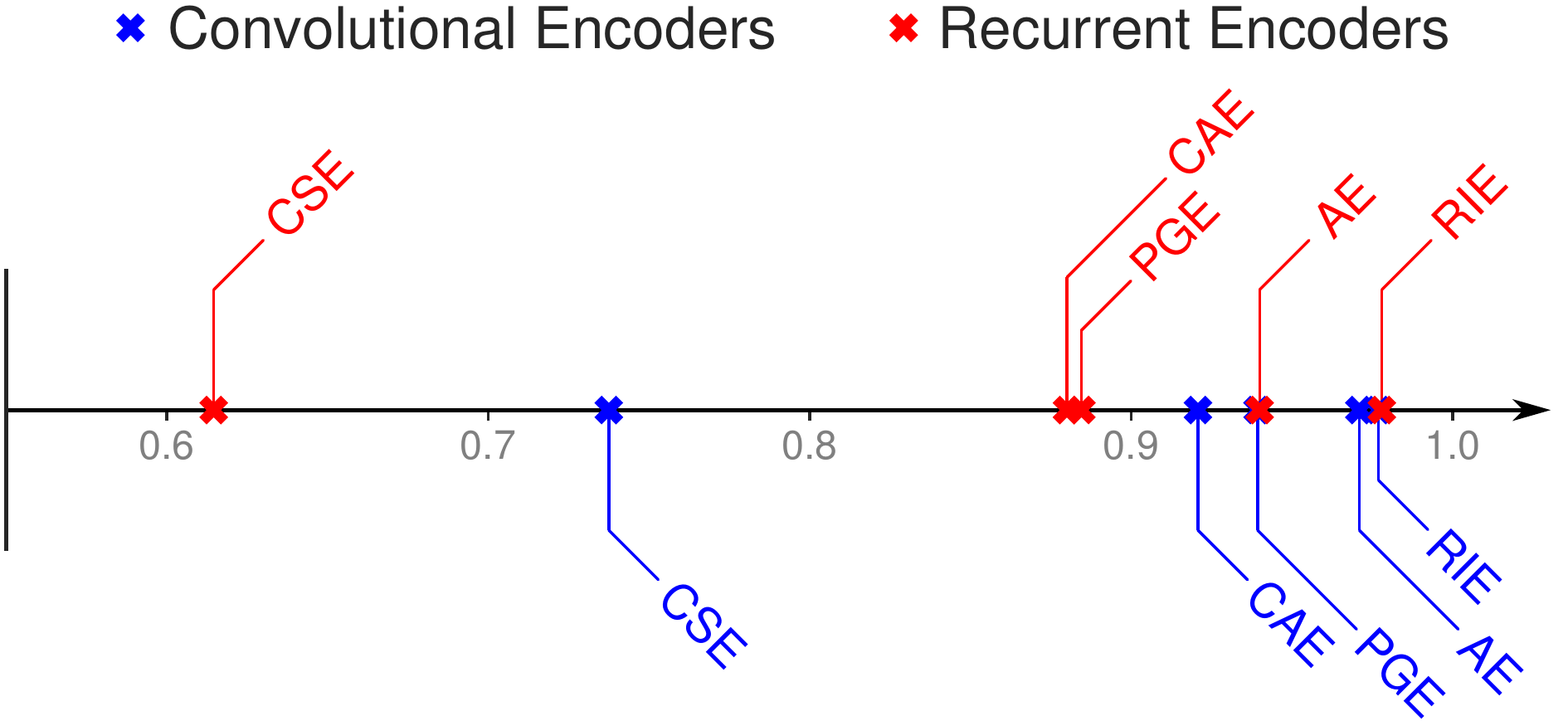}
    \caption{Network representational consistency (RC): (top) recurrent encoders and (bottom) convolutional encoders. Values closer to 1 indicates higher RC.}
    \label{fig:RC_analysis}
\end{figure}

Using CKA, we conduct pairwise similarity analysis across all six instances which yields 15 comparisons for each model type.
We report the mean of the resulting CKA values for each model type in Fig.~\ref{fig:RC_analysis}. 
First, we observe that randomly initialized encoders (RIE) are highly consistent for both architectures ($\text{mean CKA}_{RIE/RNN} \approx \text{mean CKA}_{RIE/CNN} = 0.98$). 
However, after training the encoder instances, convolutional networks are more consistent than their recurrent counterparts. 
Note that this behaviour cannot be attributed to a difference in the number of trainable parameters between the two architectures since they are comparable.    
Moreover, all decoding-based learning objectives return mean CKA values above 0.87, which indicates that their representational profiles are similar despite some noticeable differences especially among the recurrent encoders.  
The only exception to this trend are model instances trained with contrastive loss since they are significantly less consistent compared to the other learning objectives ($\text{mean CKA}_{CSE/RNN} = 0.61$ and $\text{mean CKA}_{CSE/CNN} = 0.74$). 
We emphasize that CKA is a second-order isomorphismic approach that operates on the similarity of the pairwise sample similarity matrices across different views.
Therefore, the anisotropic nature of AWEs reported in \S\ref{section_isotropy}  cannot explain their similarity-based representational profiles, and by implication, their representational consistency.

%substantial

\section{Discussion}
Acoustic word embeddings (AWEs) are vector representations that encode the sound structure and acoustic-phonetic features of spoken words.
AWEs are induced from actual acoustic realizations of speech, and therefore AWE models have to abstract away from non-linguistic dimensions of variability in speech signals (e.g., speaker characteristics, speech rate, recording conditions, etc).
While analyzing the representational geometry of semantic word embeddings is a topic that has received a substantial attention in the NLP research community, the interpretability of AWEs remains an under-explored topic and we are aware of a few prior studies in this direction \cite{matusevych2020analyzing, Abdullah2021DoAW, abdullah-etal-2021-familiar}.  
In this article, we made a number contributions in analyzing the representational geometry of AWEs and obtained research findings which we discuss and summarize in this section. \vspace{0.1cm}

\noindent
\textbf{Learning objective affects the geometry more than architecture.} \hspace{0.1cm} Our three analyses in this paper have shown that the learning objective shapes the representational geometry of the AWE encoders more than their underlying architectures. 
This finding suggests that recurrent and convolutional encoders exhibit similar inductive biases while the learning process is mainly guided by the loss function. \vspace{0.1cm}

\noindent
\textbf{AWE models tend to be anisotropic.}  \hspace{0.1cm}  Our analysis in \S\ref{section_isotropy} has shown that AWEs tend towards being minimally isotropic.
However, this behavior is not an emergent property of the training process, but rather an inherent behavior of the neural network. 
Moreover, the degree of isotropy after training the model positively correlates with the acoustic word discrimination evaluation task. 
Since different models have different degree of isotropy and the representation space is not always uniform, we conclude that any comparison between different models based on absolute distance metrics such cosine distance will definitely lead to inaccurate observations. 
\vspace{0.1cm}

\noindent
\textbf{Word distinctiveness, but not frequency, predicts category discriminability.}  \hspace{0.1cm}  
While word acoustic distinctiveness has been found in \S\ref{discriminability} to be a good predictor of the degree to which a word category is compact and well-separated in the embedding space, word frequency does not correlate with category discriminability.
In retrospective, this finding should not be surprising as frequent words tend to have shorter lengths. 
Shorter words have more phonological neighbours that are perceptually similar in form and thus they are more confusable with other words.  
Future work could employ more sophisticated linear mixed effects models to analyse the interaction between different lexical properties such as frequency, phonological neighbourhood density, and word length and their effect on word category discriminability. \vspace{0.1cm}

\noindent
\textbf{AWE models exhibit individual differences.} \hspace{0.1cm} 
Although AWE model instances trained with different random initializations are stable with respect to the performance of the evaluation task, they exhibit individual differences in their representational profiles as shown in \S\ref{representational_consistency}. 
However, the degree of the network representational consistency across different initializations depends on both the architecture and the learning objective.
Contrastive objectives are less consistent than decoding-based objectives, while recurrent encoder are less consistent than their convolutional counterparts.  \vspace{0.1cm}

\noindent
\textbf{Contrastive models have distinct representational profiles. } \hspace{0.1cm} In the analyses we presented in this paper, we observed that the contrastive encoders behave differently than other encoders trained with non-contrastive losses. 
For example, word distinctiveness has been found to be a weak predictor of category discriminability in the embedding spaces of the contrastive encoders. 
Recall that our contrastive encoders have a stronger constraint in grouping exemplars of the same category closer in the embedding space guided by the margin hyperparameter, while decoding-based model lack this constraint.  
We hypothesize that this constraint forces the models to emphasize the separability of the lexical categories in the embedding space.
Therefore, a stronger constraint seems to make contrastive encoders different compared to other learning objectives and different instances of the same contrastive encoder are less consistent in their representational geometry. 
%  even though they are trained in identical condition except their initial weights. 
% The stronger constraints, the less consistent the behavior of the encoder  compared to other models as well as instances of the same contrastive encoder trained with different random initializations.  

\section{Conclusion}
In this paper, we have taken a closer, analytical look at the representational geometry of acoustic word embeddings (AWEs) from three different, but complementary perspectives: (1) embedding space uniformity, (2) word discriminability, and (3) network representational consistency. 
We have shown that the representational spaces of AWEs tend towards being minimally isotropic, or in other words, they utilize only a few dimensions of the embedding space. 
Another finding was that most AWE models rely on word length as a feature to discriminate between lexical categories since the word discriminability index positively correlates with the number of phonemes in a word. 
Furthermore, our representational consistency analysis have shown that AWE models exhibit individual differences in their representational profiles, with the contrastive encoders being the most inconsistent across  different random initializations.  

Even though we focused on acoustic word embeddings in this paper, our analytic methodology can also be employed for the interpretability of self-supervised speech representation models such as contrastive predictive coding \cite{oord2018representation} and wav2vec \cite{schneider2019wav2vec}.
Also, the emergent representations of sublexical units such phonemes and syllables in speech neural networks can be analyzed using the our proposed methodology in this paper.

\section{Acknowledgements}
The authors would like to thank the anonymous reviewers for their encouraging feedback and insightful comments on the paper.
We express our  heartfelt thanks to Miriam Schulz for proofreading the paper. 
This research is funded by the Deutsche Forschungsgemeinschaft (DFG, German Research Foundation), Project-ID 232722074 -- SFB 1102.

%\newpage

% Entries for the entire Anthology, followed by custom entries
\bibliography{anthology,custom}

\newpage

\newpage

\appendix
\section*{Appendices}

\

\section{AWE Models}
\label{sec:models}
\begin{figure*}[t]
    \centering
    \includegraphics[width=0.9\textwidth]{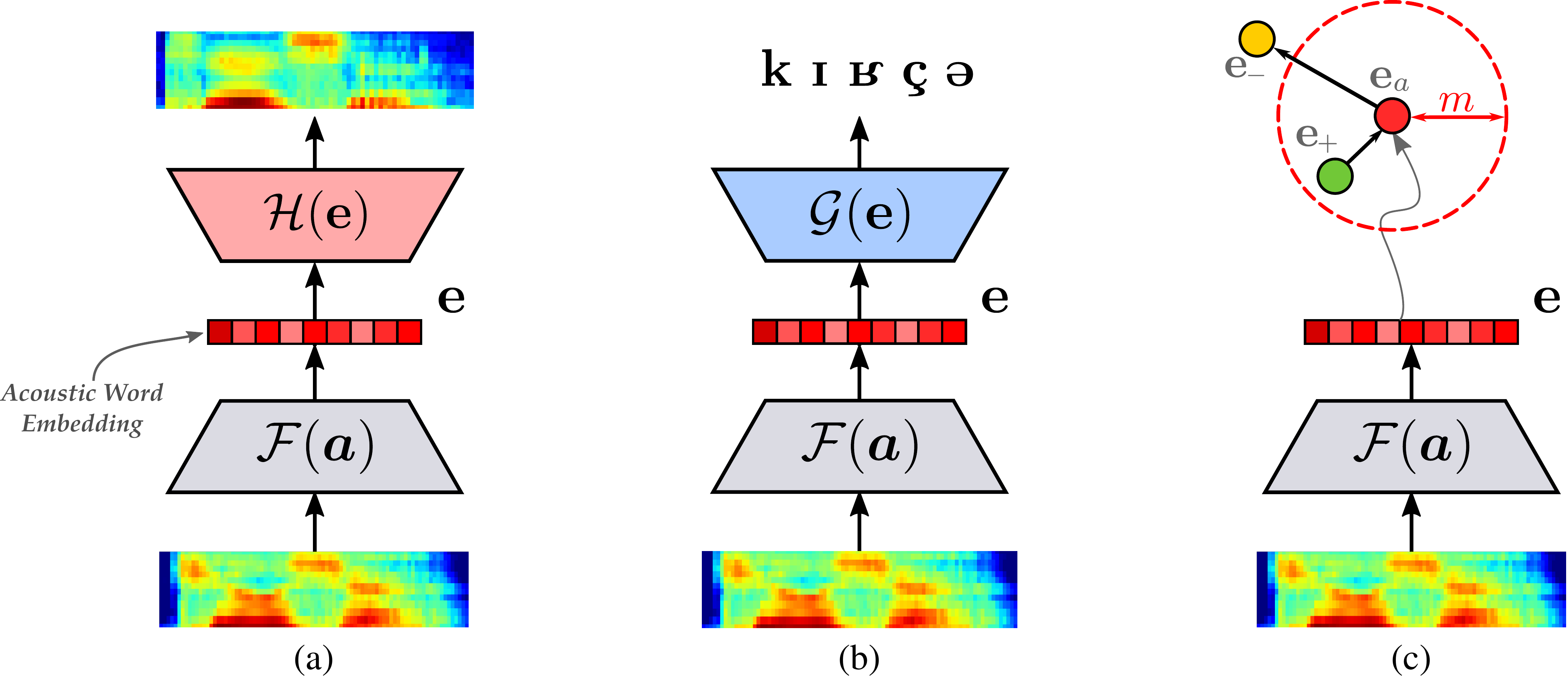}
 \caption{A visual illustration of the different learning objectives for training AWE encoders:  (a) correspondence auto-encoder (CAE): a sequence-to-sequence network with an acoustic decoder, (b) phonologically guided encoder (PGE): a sequence-to-sequence network with a phonological decoder, and (c) contrastive siamese encoder (CSE): a contrastive network trained via triplet margin loss. After training the model, only  the encoder component of the model $\mathcal{F}$ is used to produce AWEs.}
    \label{fig:all_models}
\end{figure*}

See Fig.~\ref{fig:all_models} for a visual illustration of the different learning objectives in our paper.

\section{Intrinsic Evaluation}
\begin{figure*}[t]
    \centering
    \includegraphics[width=\textwidth]{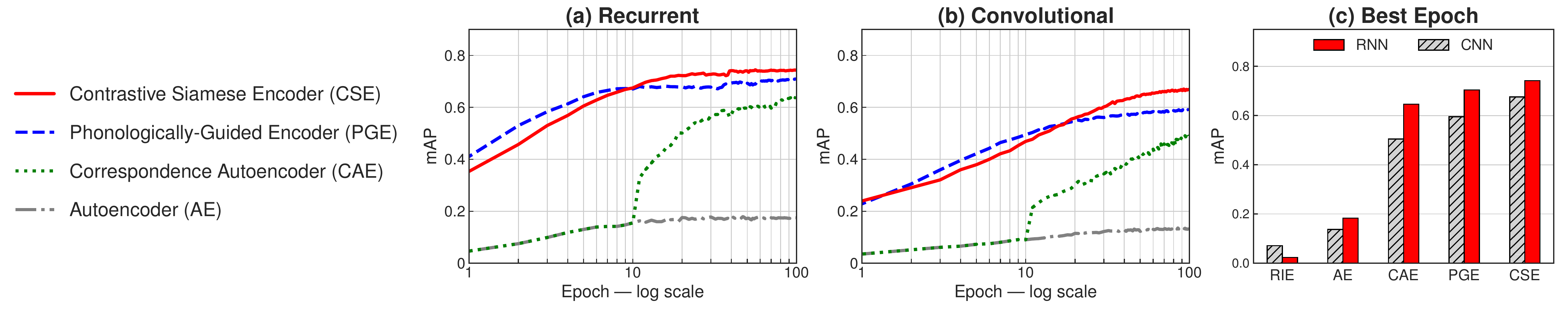}
    \caption{Evaluation on the same-different acoustic word discrimination task quantified by the word discrimination task and the mAP metric: Learning curves of 100 training epochs for  (a)  the recurrent encoder and (b) convolutional encoders.  (c) mAP of the best epoch. }
    \label{fig:AWE_evaluation}
\end{figure*}

The results of the intrinsic evaluation---same-different word discrimination task quantified by the mAP metric---is shown in Fig.~\ref{fig:AWE_evaluation}.
Note that the CAE model is pre-trained as autoencoder for 10 epochs, following prior work \cite{kamper2019truly}.

\section{Randomly Initialized CNN Encoder}
%\section*{Appendix A: Randomly Initialized Convolutional Encoder}
\label{sec:appendix_cnn_cosine}

\begin{figure}[t]
    \centering
    \includegraphics[width=0.45\textwidth]{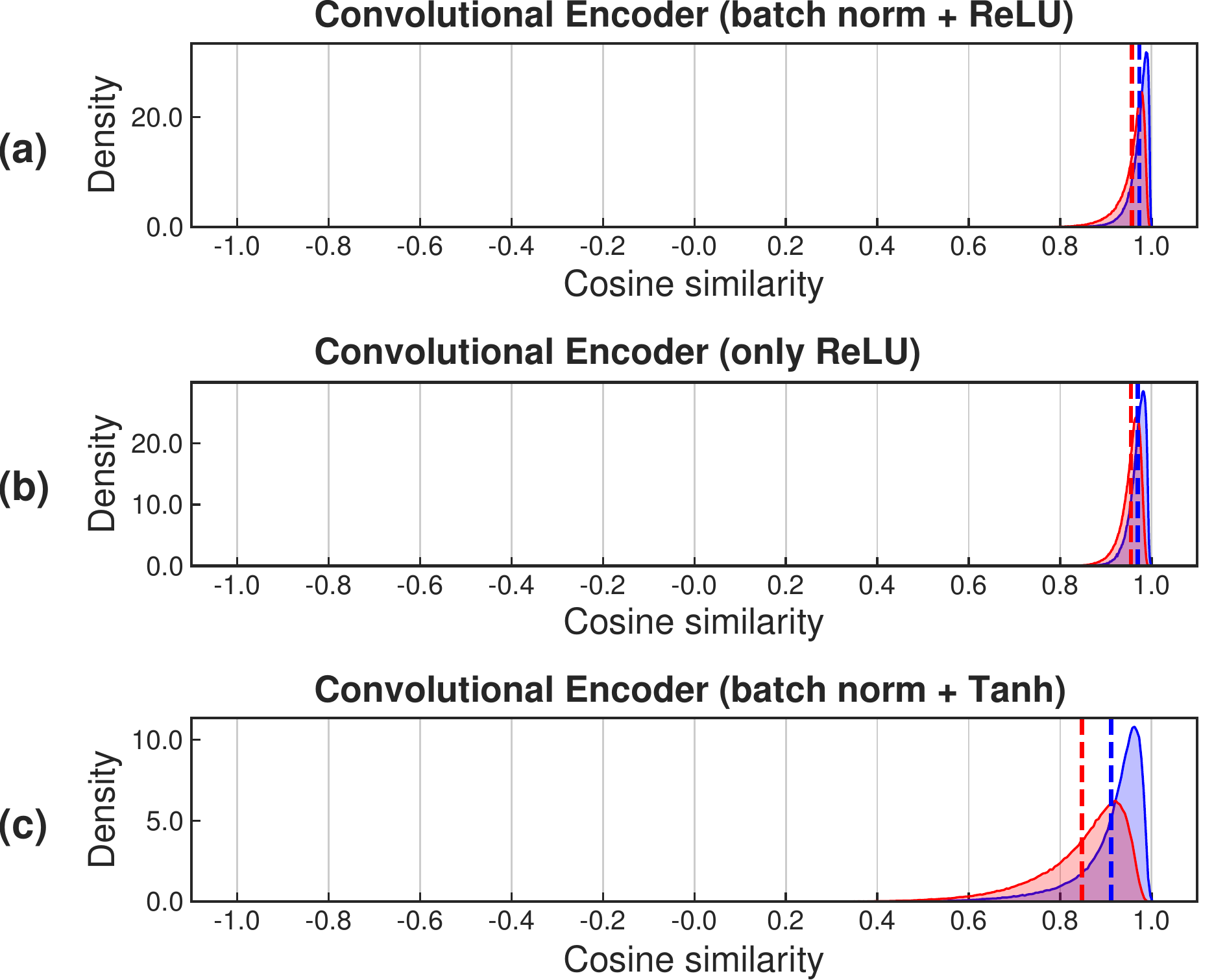}
    \caption{Cosine similarity distributions across three different variants of convolutional encoders: (a) convolutional layers with batch normalization and ReLU non-linearity, (b) convolutional layers with ReLU non-linearity but without batch normalization, and (3) convolutional layers with batch normalization and Tanh non-linearity.}
    \label{fig:CNN_cosine}
\end{figure}

To further investigate the minimally isotropic behavior and the near 1 values of cosine similarities of the untrained, randomly initialized convolutional encoder reported in \S\ref{section_isotropy}, we examine the potential contribution of two factors to this observation; batch normalization (BN) and the activation function of the convolutional layers. 
The result of this analysis is depicted in Fig.~\ref{fig:CNN_cosine}.
We observe that removing the BN layer has no effect on the distributions of cosine similarities as they remain almost identical to the encoder that has a BN layer---see Fig.~\ref{fig:CNN_cosine}(b). 
However, changing the activation function from the unbounded ReLU to the bounded Tanh in the convolutional layers makes the  distributions of cosine similarities move towards zero mean, even though they remain closer to 1 than 0. 
Therefore, this behavior seems to be related to the inner dynamics of the convolutional operation and gets amplified where the activation function in the convolutional layers are unbounded. 
Nevertheless, identifying the source of this behavior requires further investigation with different activation functions and a controlled ablation study.

\section{Performance Stability across Different Runs} 
\label{sec:appendix_RC} 
For the analysis in \S\ref{representational_consistency}, we have trained six neural network instances for each encoder type using the same training samples to investigate the performance stability and representational consistency of training runs that differ in their random seeds. 
A summary statistics for the performance on the evaluation task  measured by the mAP metric is shown in Table ~\ref{tab:mAP_performance}.
One can observe only trivial differences on the evaluation tasks.
Therefore, we conclude that the performance of different training runs is stable and our findings on the network representational consistency reported in \S\ref{representational_consistency} cannot be explained by quantitative differences, but rather by representational discrepancies due to disagreement in the geometric arrangement of the speech samples in the embedding space.

\section{Qualitative Analysis}
To further inspect the representation space and its neighborhood structure, we conduct a qualitative analysis by querying the representation space with a few word samples.
In this analysis, we compute word category centroids by averaging the word embeddings of the training samples, then we use a word centroid as a query and obtain the top-10 ranked nearest neighbors. 
The result of this analysis is shown in Fig.~\ref{tab:kNN}.
For the majority of the examples in Fig.~\ref{tab:kNN}, we observe that there is a strong word onset bias where the most similar words are those that begin with a similar sounding prefix as the query word.

\begin{table}[]
\centering
\scalebox{0.75}{
\begin{tabular}{@{}cccccc@{}}
\toprule
Objective               & Arch. & mean  & max   & min   & std    \\ \midrule
\multirow{2}{*}{AE}  & \textsc{cnn}    & 0.137 & 0.141 & 0.133 & 0.0026 \\
                     & \textsc{rnn}    & 0.183 & 0.186 & 0.179 & 0.0024 \\ \midrule
\multirow{2}{*}{CAE} & \textsc{cnn}    & 0.505 & 0.510  & 0.500 & 0.0040 \\
                     & \textsc{rnn}    & 0.646 & 0.650  & 0.643 & 0.0029 \\ \midrule
\multirow{2}{*}{PGE} & \textsc{cnn}    & 0.595 & 0.599 & 0.592 & 0.0033 \\
                     & \textsc{rnn}    & 0.704 & 0.710  & 0.687 & 0.1000 \\ \midrule
\multirow{2}{*}{CSE} & \textsc{cnn}    & 0.676 & 0.680  & 0.674 & 0.0023 \\
                     & \textsc{rnn}    & 0.742 & 0.745 & 0.739 & 0.0027 \\ \bottomrule
\end{tabular}
}
\caption{mAP statistics across six different runs for each model type. }
\label{tab:mAP_performance}
\end{table}

% Please add the following required packages to your document preamble:
% \usepackage{booktabs}
% \usepackage{multirow}
\begin{table*}[]
\centering
\scalebox{0.70}{
\begin{tabular}{@{}l|llll|llll@{}}
\toprule
\multirow{2}{*}{\textbf{Query} ($\downarrow$)}                            & \multicolumn{4}{l}{Convolutional Encoders (CNN)}                                    & \multicolumn{4}{|l}{Recurrent Encoders (RNN)}                                     \\ \cmidrule(l){2-9} 
                               & AE          & CAE           & PGE           & CSE          & AE          & CAE           & PGE            & CSE          \\ \midrule
\multirow{10}{*}{\textbf{mentioned}}    & mention     & mention       & mention       & mention      & mention     & mention       & mention        & mention      \\
                               & wretched    & mansion       & mansion       & mansion      & wretched    & mansion       & mansion        & mansion      \\
                               & nation      & motion        & legends       & merchant     & nation      & merchant      & merchant       & merchant     \\
                               & midst       & merchant      & management    & mission      & merchant    & motion        & legends        & mission      \\
                               & merchants   & making        & merchants     & mental       & motion      & nation        & mountain       & pinching     \\
                               & motion      & wilson        & magic         & vincent      & merchants   & making        & merchants      & massive      \\
                               & merchant    & nation        & matrons       & pinching     & midst       & vincent       & mission        & mental       \\
                               & message     & midst         & mission       & medicine     & milking     & nineteen      & magician       & transient    \\
                               & regiment    & missing       & merchant      & crouching    & winter      & nature        & motion         & motion       \\
                               & winter      & nature        & magician      & midst        & vessel      & rachel        & wretched       & hudson       \\ \midrule
\multirow{10}{*}{\textbf{intellectual}} & individual  & introduction  & intellect     & intellect    & individual  & individual    & individual     & introduction \\
                               & interesting & individual    & individual    & adjoining    & interesting & introduction  & intelligence   & individual   \\
                               & indifferent & interrupted   & introduction  & recollection & neglected   & uncomfortable & introduction   & immature     \\
                               & newton      & indifference  & intelligent   & delightful   & petition    & intelligent   & intellect      & objection    \\
                               & institution & attraction    & encouragement & individual   & magician    & intelligence  & uncomfortable  & implacable   \\
                               & departure   & intellect     & interrupted   & employing    & hokosa      & interesting   & intelligent    & delightful   \\
                               & imitation   & immature      & intelligence  & impetuous    & compassion  & invisible     & interpretation & theatrical   \\
                               & hokosa      & indifferent   & indifferently & employed     & departure   & interrupted   & industrial     & thoughtful   \\
                               & encountered & encouragingly & unconditional & natural      & convention  & imperfectly   & incapable      & industrial   \\
                               & neglected   & implacable    & impetuous     & accumulated  & consulted   & incredible    & insensible     & election     \\  \midrule
\multirow{10}{*}{\textbf{maker}}        & labor       & naked         & naked         & baker        &  labor       & nature        & baker          & liquor       \\
                               & liquor      & natured       & liquor        & naked        & nature      & local         & nature         & negro        \\
                               & labored     & nature        & natured       & negro        & walker      & naked         & liquor         & eaten        \\
                               & labour      & local         & nature        & liquor       & local       & labour        & labour         & baker        \\
                               & wicker      & labour        & baker         & local        & naked       & labor         & labor          & nature       \\
                               & leaping     & major         & major         & native       & labour      & major         & major          & labor        \\
                               & lifted      & labor         & negro         & nature       & rachel      & natured       & negro          & naked        \\
                               & walker      & native        & native        & major        & liquor      & baker         & neighbors      & newspaper    \\
                               & local       & making        & wicker        & matrons      & labored     & liquor        & vapor          & mink         \\
                               & nature      & navy          & labor         & vigor        & leaping     & negro         & labors         & vigour       \\ \midrule
\multirow{10}{*}{\textbf{profession}}   & position    & procession    & procession    & professor    & position    & procession    & procession     & professor    \\
                               & proceed     & professor     & professor     & sufficient   & professors  & possession    & proportion     & procession   \\
                               & positions   & position      & position      & procession   & possessions & position      & perfection     & perfection   \\
                               & physician   & possession    & petition      & professors   & proceeded   & professor     & possession     & sufficient   \\
                               & proceeded   & professors    & pushing       & efficiency   & physician   & possessions   & protection     & proposition  \\
                               & possessed   & pushing       & professors    & efficient    & condition   & permission    & proportions    & proportion   \\
                               & prison      & perfection    & possession    & petition     & procession  & discussion    & position       & production   \\
                               & possessions & positions     & physician     & prevent      & presumption & positions     & possessions    & petition     \\
                               & perfect     & discussion    & positions     & position     & protested   & commission    & professor      & compassion   \\
                               & discussion  & preferred     & precious      & physician    & proceed     & physician     & petition       & pushing   \\ \midrule  
 \multirow{10}{*}{\textbf{seized}}       & ceased      & ceased        & ceased        & thieves      & ceased      & ceased        & ceased         & thieves      \\
                               & freedom     & season        & seizing       & ceased       & faded       & feasts        & thieves        & ceased       \\
                               & seated      & thieves       & season        & season       & cities      & scenes        & saves          & fuse         \\
                               & faded       & saves         & thieves       & feast        & singing     & thieves       & seats          & jesus        \\
                               & singing     & seems         & saves         & seizing      & scenes      & saves         & seems          & spheres      \\
                               & scenes      & scenes        & ceasing       & feared       & feeding     & seems         & scenes         & feels        \\
                               & season      & ceasing       & seems         & ceasing      & season      & feast         & seemed         & cities       \\
                               & cities      & saints        & feast         & saves        & sweetest    & saints        & feast          & season       \\
                               & field       & feast         & seats         & species      & seated      & faced         & saved          & seats        \\
                               & seeming     & sins          & seemed        & speed        & saying      & seemed        & seizing        & scenes      \\ \midrule   
                               
\multirow{10}{*}{\textbf{experiments}}  & experiment   & experiment    & experiment    & experiment    & experiment   & experiment    & experiment     & experiment   \\
                               & experience   & experienced   & experience    & experienced   & experienced  & experienced   & experienced    & attendants   \\
                               & experienced  & experience    & experienced   & garments      & experience   & experience    & experience     & extremities  \\
                               & experiences  & experiences   & experiences   & extermination & extinguished & experiences   & experiences    & islands      \\
                               & extinguished & extremities   & expense       & expense       & experiences  & exposed       & expressions    & experienced  \\
                               & exchange     & established   & embarrassment & experience    & expected     & extremities   & extermination  & prominence   \\
                               & extremities  & extraordinary & expanse       & aramis        & exchange     & expense       & extremities    & edmunds      \\
                               & expressions  & extinguished  & extraordinary & disturbance   & expressions  & expanse       & extremity      & instruments  \\
                               & extremely    & extremity     & extremities   & examined      & extremities  & extinguished  & expression     & attendance   \\
                               & extremity    & expanse       & expressions   & vanished      & extent       & exclusion     & expensive      & commons    \\ \midrule     
\end{tabular}
}
\caption{Top-10 nearest word embedding centroids for a word sample. }
\label{tab:kNN}
\end{table*}

\end{document}